\documentclass{article}

    \PassOptionsToPackage{numbers, compress}{natbib}
\usepackage[preprint]{neurips_2026}

\usepackage[utf8]{inputenc} % allow utf-8 input
\usepackage[T1]{fontenc}    % use 8-bit T1 fonts
\usepackage{hyperref}       % hyperlinks
\usepackage{url}            % simple URL typesetting
\usepackage{booktabs}       % professional-quality tables
\usepackage{amsfonts}       % blackboard math symbols
\usepackage{nicefrac}       % compact symbols for 1/2, etc.
\usepackage{microtype}      % microtypography
\usepackage{xcolor}         % colors
\usepackage{multirow}
\usepackage{amsmath} 
\usepackage{graphicx}
\usepackage{wrapfig}
\usepackage{amssymb}
\usepackage{enumitem}

\title{Bringing Multimodal Large Language Models to Infrared-Visible Image Fusion Quality Assessment}

\author{
\hspace*{-0.3cm}
  Yuchen Guo$^{1,}$\thanks{Equal contribution. \\ Correspondence to: yuchenguo2027@u.northwestern.edu, wfsu@bnbu.edu.cn.} \quad
  Junli Gong$^{2,}$\footnotemark[1] \quad
  Yao Lu$^{3}$ \quad
  Xintong Xu$^{5}$ \quad
  Yiuming Cheung$^{4}$ \quad
  Weifeng Su$^{5}$ \\ \\
  $^1$Northwestern University \quad
  $^2$Northeastern University \quad
  $^3$University of Washington \\
  $^4$Hong Kong Baptist University \quad
  $^5$Beijing Normal - Hong Kong Baptist University
}

\begin{document}

\maketitle

\begin{abstract}
Infrared-Visible image fusion (IVIF) aims to integrate thermal information and detailed spatial structures into a single fused image to enhance perception. However, existing evaluation approaches tend to over-optimize both hand-crafted no-reference statistics and full-reference metrics that treat the source images as pseudo ground truths. Recent IVIF 
reward-modelling efforts learn from human ratings but use scalar regression on aggregated scores, neither leveraging the reasoning of Multimodal Large Language Models (MLLMs) nor encoding per-image perceptual ambiguity in their supervision, but naively introducing MLLMs with discrete one-hot supervision likewise collapses fused images of similar quality into different rating levels. To address this, we introduce \textbf{\textit{FuScore}}, which utilizes an MLLM to mimic human visual perception by producing \textit{continuous} quality score, rather than discrete level predictions, enabling fine-grained discrimination among fused images of similar quality. We exploit the agreement among four IVIF-specific sub-dimensions to construct a per-image soft label whose sharpness reflects how consensual the overall judgment is. We further introduce a tripartite objective combining per-image distributional supervision, within-source-pair Thurstone fidelity for method-level ordering, and cross-source-pair Thurstone fidelity for scene-level ordering across scenes. Extensive experiments demonstrate that FuScore achieves state-of-the-art correlation with human visual preferences.
\end{abstract}

% FuScore also generates Chain-of-Thought (CoT) reasoning rationales, providing interpretable linguistic justifications for its perceptual assessment.

\section{Introduction}

Infrared-Visible image fusion (IVIF) is an important technique in image processing, which aims to combine both thermal and structural information from two source images into a single fused image \cite{zhang2023visible,wang2005comparative,li2017pixel}. By integrating complementary thermal and structural cues into a unified image representation, IVIF has been widely applied in a variety of real-world scenarios, including security surveillance, military target detection, scene understanding, and assisted driving. Existing IVIF models can be broadly categorized into two approaches: discriminative methods such as carefully designed CNN-based and Transformer-based models \cite{zhang2020ifcnn, ma2022swinfusion}; generative approaches, including autoencoder-based methods \cite{guo2025dae, zhao2023cddfuse}, GAN-based \cite{liu2022target, ma2020ddcgan} and diffusion-based methods \cite{tang2025mask, zhao2023ddfm}. 

\begin{figure*}[t]
\centering
\includegraphics[width=\linewidth]{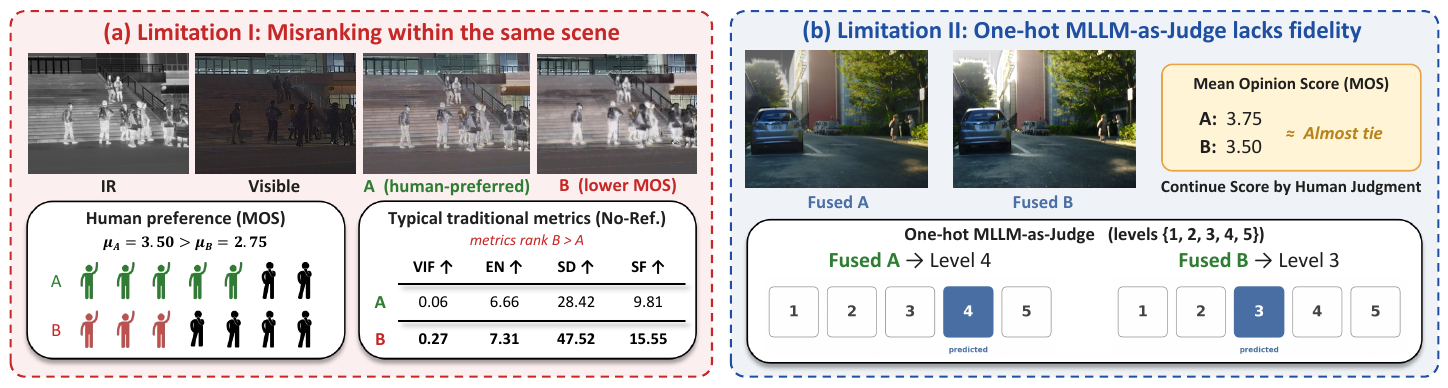}
\vspace{-12pt}
\caption{Two limitations of current IVIF quality assessment.
\textbf{(a) Misranking within the same scene:} for a single
source pair, traditional no-reference metrics (VIF, EN, SD,
SF) rank the human-disfavoured fused image B above the
human-preferred A (MOS $\mu_A{=}3.50 > \mu_B{=}2.75$).
\textbf{(b) One-hot MLLM-as-Judge lacks fidelity:} two fused
images with near-tie human MOS ($3.75$ vs.\ $3.50$) are
forced into distinct discrete levels ($4$ vs.\ $3$) by an
MLLM trained with one-hot supervision over five quality
levels.}
\label{fig:intro}
\vspace{-9pt}
\end{figure*}

With the rapid advancement of deep learning, above IVIF models have achieved substantial progress in improving fusion quality. However, a fundamental limitation remains: these existing methods are still predominantly optimized for hand-crafted traditional no-reference statistics such as Entropy (EN) \cite{roberts2008assessment} and Mutual Information (MI) \cite{eskicioglu1995image}, or toward full-reference objectives that treat source image pairs as pseudo ground truths, such as Structural Similarity Index (SSIM) \cite{wang2004image} and Peak Signal-to-Noise Ratio (PSNR) \cite{hore2010image}, due to the absence of authentic ground-truth fused images in IVIF. As a result, these objectives often fail to faithfully reflect human experts perceptual preference, and may even assign higher scores to a visually inferior fused image (Fig.~\ref{fig:intro}).

A recent IVIF reward-modelling effort, EVAFusion 
\cite{liu2026bridging}, addresses the human-alignment gap by 
training a ViT-based regressor with MSE on aggregated overall 
scores. Yet it neither leverages the linguistic reasoning of 
MLLMs nor encodes how \emph{consensual} each image's mean 
score is across the underlying multi-criterion judgments. Another 
naive remedy of introducing MLLMs with the discrete one-hot 
supervision used in prior MLLM-based IQA~\cite{wu2023q} is unable to capture subtle perceptual differences and carries no signal about per-image annotation consensus either. This leaves a fundamental question largely unanswered: \textbf{\textit{how can MLLMs produce continuous IVIF quality scores under
supervision that encodes per-image annotation consensus?}}

We answer this question with \textbf{\textit{FuScore}}, an 
MLLM-based scorer that yields a continuous quality score for 
each fused image rather than forcing it into a single discrete 
level. Specifically, instead of training the MLLM to emit a 
single quality token, we read out the decoding probabilities 
over the predefined quality-level tokens and supervise this 
distribution with a soft label, so that the final score, 
computed as the expectation under the predicted distribution, 
can take any real value within the rating range and capture 
subtle quality differences that one-hot supervision collapses. 
A key challenge, however, lies in how to construct the soft 
label itself: naively smoothing a discrete overall score with a 
fixed-width Gaussian ignores the fact that fused images vary 
substantially in how consensual their perceptual quality is. 
To address this, we exploit the structure of existing IVIF 
annotations, where each fused image is labeled along four 
sub-dimensions: thermal retention, texture preservation, 
artifacts, and sharpness, and use the agreement among these 
sub-dimensions as a per-image consensus signal. When the four 
criteria agree, the soft label is sharply concentrated around 
the overall score; when they disagree, the label is broadened 
to reflect that the overall judgment is less consensual across 
reasonable aggregation rules. We further observe that 
subjective fusion quality exhibits both within-source-pair 
variation, which captures how different fusion methods perform 
on the same scene, and cross-source-pair variation, which 
reflects scene-dependent difficulty. Since neither per-image 
supervision nor within-scene comparisons alone can cover both 
variation components, FuScore adopts a tripartite objective 
that jointly enforces per-image distributional supervision, 
within-source-pair Thurstone fidelity for method-level ordering, 
and cross-source-pair Thurstone fidelity for scene-level 
ordering, with the per-pair margin reusing the per-image label 
widths derived from sub-dimension agreement. Our core 
contributions are summarized as:

\begin{itemize}[itemsep=1pt, topsep=1pt]
\item We propose FuScore, the first MLLM-based IVIF quality 
scorer trained with distribution-level supervision, generating continuous score of fused images.
\item We introduce a sub-dimension-aware soft label construction 
that modulates label sharpness by the agreement among four 
IVIF-specific sub-dimensions, allowing the supervision target to 
faithfully reflect how consensual the overall perceptual 
judgment is on each fused image.
\item We design a tripartite training objective that jointly 
enforces per-image distributional supervision, within-source-pair 
Thurstone fidelity for method-level ordering, and 
cross-source-pair Thurstone fidelity for scene-level ordering, 
with the per-pair margin reusing the per-image label widths 
derived from sub-dimension agreement.
\item On the EVAFusion benchmark and a 5-expert Semantic RT 
pilot, FuScore achieves state-of-the-art correlation with human 
visual preferences across all reported rank metrics, and 
empirically tracks both data-side ranking difficulty and 
human-side inter-rater disagreement.
\end{itemize}

\section{Related Works}
\textbf{Infrared-Visible Image Fusion.}
Infrared-visible image fusion (IVIF) is a crucial field of 
low-level computer vision. Early CNN-based encoder-decoder 
architectures such as IFCNN~\cite{zhang2020ifcnn} rely on 
hand-designed fusion rules, while Transformer-based models like 
SwinFusion~\cite{ma2022swinfusion} capture long-range 
cross-modality dependencies via self-attention. Generative 
approaches instead model the data distribution of fused images: 
autoencoders~\cite{zhao2023cddfuse, guo2025dae, guo2024fuse4seg} 
learn modality-specific representations, 
GANs~\cite{ma2020ddcgan, liu2022target} encourage perceptual 
realism via adversarial training, and 
diffusion~\cite{zhao2023ddfm, tang2025mask} formulates fusion as 
conditional generation. Despite diverse architectures, all 
optimize either no-reference statistics or pseudo-reference 
objectives derived from source pairs, both poorly correlated 
with human perception, motivating a perceptually aligned 
quality scorer.

\textbf{Image Quality Assessment.}
IQA methods~\cite{wang2004image, wang2006modern, yang2022maniqa, 
wang2023exploring, ke2021musiq} predict a per-image quality 
score under no-reference or full-reference paradigms. MLLMs have 
recently been adapted as perceptual scorers on natural images: 
Q-Align~\cite{wu2023q} fine-tunes an MLLM to predict five 
discrete quality levels via level-token probabilities, and 
DeQA~\cite{you2025teaching} replaces one-hot supervision with 
soft-label distributions from inter-rater variance. In IVIF, 
fusion quality is still commonly evaluated using handcrafted 
metrics~\cite{cheng2026evanet}, no-reference measures such as 
EN~\cite{roberts2008assessment}, MI~\cite{eskicioglu1995image}, 
and VIF~\cite{ma2019infrared}, or full-reference 
SSIM~\cite{wang2004image} and PSNR~\cite{hore2010image} treating 
source pairs as pseudo references. Recent IVIF reward-modelling efforts~\cite{liu2026bridging, chang2023semantic} 
trains a ViT-based regressor with MSE on aggregated overall 
scores; it is, however, neither MLLM-based nor 
distribution-supervised, and its scalar regression target 
provides no signal about per-image perceptual consensus. 
MLLM-based perceptual scorers and per-image ambiguity 
modelling have not yet been explored for IVIF.

\section{Methodology}
\label{sec:method}

We present \textbf{FuScore}, an MLLM-based scorer that assigns each fused 
image a continuous quality score reflecting human perceptual preference. 
We first formulate the scoring problem and the MLLM output formulation 
(Sec.~\ref{sec:method:setup}), then introduce our sub-dimension-aware soft 
label construction (Sec.~\ref{sec:method:softlabel}), and finally describe 
the tripartite training objective that jointly supervises per-image 
distributional matching, within-source-pair fidelity, and cross-source-pair 
ranking (Sec.~\ref{sec:method:objective}).

\subsection{Problem Setup}
\label{sec:method:setup}

\noindent\textbf{Input and supervision.}  
We leverage EVAFusion~\cite{liu2026bridging}, a large-scale IVIF 
quality benchmark comprising $9{,}350$ fused images produced by $11$ 
representative IVIF methods over $850$ source image pairs. For each 
fused image $x_i$, EVAFusion provides aggregated human ratings along 
four IVIF-specific sub-dimensions $\{s_i^{(k)}\}_{k=1}^{4}$ (thermal 
retention, texture preservation, artifacts, and sharpness), together 
with an overall quality score $y_i$, all in $[1,5]$. The model input is source-conditioned and full-reference:
\begin{equation}
\mathbf{x}_i = (\mathrm{IR}_i,\ \mathrm{VIS}_i,\ \mathrm{Fused}_i),
\end{equation}
so that the scorer evaluates the fused image jointly with its two source 
modalities.

\noindent\textbf{From discrete tokens to a continuous score.} 
We instantiate FuScore on a pretrained MLLM 
$f_\theta$. Given the input $\mathbf{x}_i$ and a quality-query prompt, 
$f_\theta$ produces logits $z_i^{(l)}$ over five predefined quality-level 
tokens $\mathcal{L}=\{1,2,3,4,5\}$ at the decoding step. Applying softmax 
over these five tokens yields a categorical distribution
\begin{equation}
q_i(l\,|\,\theta) = \frac{\exp(z_i^{(l)})}{\sum_{l'\in\mathcal{L}} \exp(z_i^{(l')})}, 
\qquad l\in\mathcal{L}.
\label{eq:softmax}
\end{equation}
The final continuous score is then read out as the expectation of the 
quality level under this distribution:
\begin{equation}
\hat{y}_i = \mathbb{E}_{q_i}[L] = \sum_{l\in\mathcal{L}} l\cdot q_i(l).
\label{eq:readout}
\end{equation}
This expectation lies in $[1,5]$ and varies continuously with the predicted
distribution, so that perceptually similar fused images can be assigned
distinct fine-grained scores rather than being collapsed into the same
discrete level. Importantly, the model emits a single distribution $q_i$
rather than separately predicting a variance term: all uncertainty
modeling thus enters via the supervision target's width (described next),
while predictive uncertainty at inference is read off as the standard
deviation of $q_i$ and used in Sec.~\ref{sec:ablation}.

\subsection{Sub-Dimension-Aware Soft Label Construction}
\label{sec:method:softlabel}

To supervise the predicted distribution $q_i$, we construct a per-image 
soft label $p_i$ over the five quality levels. The center of $p_i$ is 
determined by the overall score $y_i$, while its sharpness is modulated 
by a per-image consensus signal derived from the sub-dimension annotations. 

\noindent\textbf{Sub-dimension agreement signal.} 
A fused image is uniformly easy to judge when all four sub-dimension 
scores agree: thermal, texture, artifact, and sharpness ratings all 
point to the same quality level---and is intrinsically more arguable when 
they disagree, since different reasonable weightings of these criteria 
would lead to different overall judgments. We capture this property with 
the empirical standard deviation of the four sub-dimension scores:
\begin{equation}
\delta_i \;\triangleq\; \mathrm{std}\!\left(s_i^{(1)},\, s_i^{(2)},\, 
s_i^{(3)},\, s_i^{(4)}\right).
\label{eq:delta}
\end{equation}
A small $\delta_i$ indicates a consensual overall judgment; a large 
$\delta_i$ indicates that the four perceptual criteria disagree, so 
that the overall score is less consensual across reasonable aggregation 
rules. We empirically validate that $\delta_i$ tracks genuine 
annotation ambiguity, rather than merely reflecting multi-attribute 
trade-off, in Sec.~\ref{sec:exp:delta_validation}.

\noindent\textbf{Per-image label width.}
Following the Thurstone framework~\cite{thurstone2017law}, we model 
each fused image's latent perceptual quality as a Gaussian on the 
rating scale, parameterized by a per-image center $\mu_i$ and width 
$\sigma_i$. Setting $\mu_i = y_i$ anchors the label at the overall 
score; the width $\sigma_i$ is then defined to scale linearly with 
the sub-dimension agreement signal $\delta_i$ on top of an irreducible 
noise floor $\sigma_0$:
\begin{equation}
\sigma_i \;=\; \mathrm{clip}\!\left(\sigma_0 + \lambda_c\,\delta_i,\;
\sigma_{\min},\,\sigma_{\max}\right).
\label{eq:sigma}
\end{equation}
The floor $\sigma_0$ accounts for residual annotation noise even when
all four criteria agree (e.g.,\ from discretizing continuous perceptual
judgments into the integer scale $\{1,\dots,5\}$); $\lambda_c$ controls
how sensitively the label width responds to sub-dimension disagreement;
and the soft clip $\sigma_i \in [\sigma_{\min}, \sigma_{\max}]$
prevents degenerate distributions at the extremes.

\noindent\textbf{Gaussian binning with first-moment preservation.} 
Given the center $\mu_i = y_i$ and width $\sigma_i$, we discretize a 
Gaussian density over the five quality levels:
\begin{equation}
p_i(l) \;\propto\; \mathcal{N}\!\left(l;\, \mu_i,\, \sigma_i^2\right), 
\qquad l\in\mathcal{L}.
\label{eq:plabel}
\end{equation}
Following DeQA \cite{you2025teaching}, we further post-adjust $p_i$ so that its 
first moment exactly recovers the overall score:
\begin{equation}
\sum_{l\in\mathcal{L}} p_i(l) = 1, 
\qquad 
\sum_{l\in\mathcal{L}} l\cdot p_i(l) = \mu_i.
\label{eq:postadjust}
\end{equation}
This first-moment preservation is essential: it ensures that the
distributional target $p_i$ and the scalar target $\mu_i$ are mutually
consistent, so that minimizing the divergence between $q_i$ and $p_i$
also drives the predicted score $\hat{y}_i = \mathbb{E}_{q_i}[L]$ toward
$\mu_i$. As a result, the per-image soft label is sharply concentrated
around $\mu_i$ when the sub-dimensions agree, and broadens around $\mu_i$
when they conflict, faithfully reflecting how consensual the overall
judgment is on each fused image.

\subsection{Tripartite Training Objective}
\label{sec:method:objective}

A natural choice would be to train FuScore by matching $q_i$ to $p_i$ 
alone. However, the primary metrics by which IQA models are evaluated
(\textit{e.g.}, SRCC and KRCC) are non-differentiable
functions of the predicted ranking; per-image distributional alignment
only supervises ranking implicitly through each image's expected score
and provides a weak signal in regimes where the GT $\mu$-gap between
two images is small relative to label noise. We therefore add two
differentiable Thurstone-style \cite{thurstone2017law} ranking proxies whose scopes mirror the two ranking metrics that the field reports.

\begin{figure*}[t]
\centering
\includegraphics[width=\linewidth]{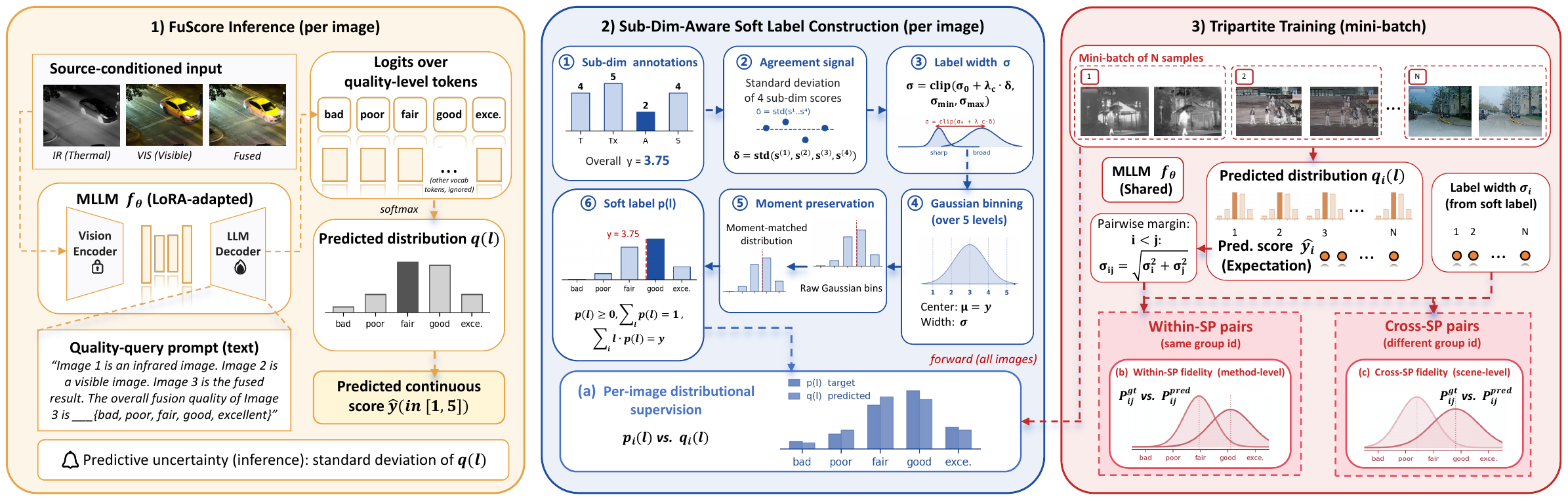}
\vspace{-12pt}
\caption{Framework of our FuScore. a) FuScore Inference for per image. b) Sub-dim-aware soft label construction. c) Tripartite training process.}
\label{fig:method}
\vspace{-12pt}
\end{figure*}

\noindent\textbf{Within- and cross-source-pair scopes match the
evaluation metrics.}
The total variance of $y_i$ admits a natural decomposition by source
pair $g$:
\begin{equation}
\mathrm{Var}(y) \;=\;
\underbrace{\mathbb{E}_{g}\!\left[\mathrm{Var}(y\,|\,g)\right]}_{\text{within-source-pair}}
\;+\;
\underbrace{\mathrm{Var}_{g}\!\left[\mathbb{E}(y\,|\,g)\right]}_{\text{cross-source-pair}}.
\label{eq:vardecomp}
\end{equation}
The within-source-pair component captures how different fusion methods
perform on the same scene, the quantity that per-source-pair Kendall
$\tau$ measures at evaluation time, while the cross-source-pair
component captures how scene-level difficulty itself shifts the
achievable quality, which is what pooled SRCC reflects. We therefore train FuScore with three complementary terms whose summation scopes correspond to these two evaluation regimes plus the per-image shape signal.

\noindent\textbf{Per-image distributional supervision $\mathcal{L}_{\mathrm{KL}}$.} 
The first term aligns the predicted distribution $q_i$ with the soft 
label $p_i$ via Kullback--Leibler divergence:
\begin{equation}
\mathcal{L}_{\mathrm{KL}} \;=\; 
\frac{1}{B}\sum_{i=1}^{B} \mathrm{KL}\!\left(p_i \,\big\|\, q_i\right),
\label{eq:lkl}
\end{equation}
where $B$ is the batch size. By the first-moment preservation in 
Eq.~\eqref{eq:postadjust}, this term simultaneously supervises both the 
shape of the level distribution and the location of the predicted 
expectation $\hat{y}_i$.

\noindent\textbf{Within-source-pair fidelity $\mathcal{L}_{\mathrm{fid}}$.} 
Within a source pair, ``which fused image is better'' is a semantically 
well-defined comparison, since both fused images depict the same scene 
and arise from the same content-preservation task. We therefore impose 
a Thurstone-style pairwise fidelity loss on every pair $(i,j)$ that
shares the same source pair $g(i)=g(j)$. We use a single per-pair
margin
\begin{equation}
\sigma_{ij} \;\triangleq\; \sqrt{\sigma_i^2 + \sigma_j^2},
\label{eq:sigmapair}
\end{equation}
inherited from the soft-label widths of Eq.~\eqref{eq:sigma}, and
define the Ground Truth (GT) and predicted pairwise preference
probabilities as
\begin{equation}
P_{ij}^{\mathrm{gt}} = \Phi\!\left(\frac{\mu_i - \mu_j}{\sigma_{ij}}\right), \qquad
P_{ij}^{\mathrm{pred}} = \Phi\!\left(\frac{\hat{y}_i - \hat{y}_j}{\sigma_{ij}}\right),
\label{eq:thurstone_probs}
\end{equation}
where $\Phi$ is the standard Gaussian CDF. We score the two
distributions with the DeQA-Score fidelity loss
\cite{you2025teaching}, which is bounded in $[0,1]$ per pair and
self-saturates at confident pairs:
\begin{equation}
\mathcal{L}_{\mathrm{fid}} \;=\; \frac{1}{|\mathcal{S}_{\mathrm{fid}}|}\!\!\
\sum_{(i,j)\in\mathcal{S}_{\mathrm{fid}}}\!\!\!
\Big[ 1 - \sqrt{P_{ij}^{\mathrm{gt}} P_{ij}^{\mathrm{pred}}}
- \sqrt{\big(1{-}P_{ij}^{\mathrm{gt}}\big)\big(1{-}P_{ij}^{\mathrm{pred}}\big)} \ \Big],
\label{eq:lfid}
\end{equation}
where $\mathcal{S}_{\mathrm{fid}} = \{(i,j):\, g(i)=g(j),\, i<j\}$
indexes the within-source-pair pairs in the batch.
The pairwise margin $\sigma_{ij}$ widens when either image has a less
consensual overall judgment, so that high-$\delta$ pairs contribute
looser comparisons and low-$\delta$ pairs contribute sharper
ones---an uncertainty-aware reweighting that follows naturally from the 
soft-label construction in Sec.~\ref{sec:method:softlabel}.

\noindent\textbf{Cross-source-pair Thurstone fidelity
$\mathcal{L}_{\mathrm{xfid}}$.}
The within-source-pair fidelity $\mathcal{L}_{\mathrm{fid}}$ is restricted
to comparisons that share a scene; the cross-source-pair component of
Eq.~\eqref{eq:vardecomp} -- the variation in achievable quality across
\emph{different} scenes -- is left unsupervised by
$\mathcal{L}_{\mathrm{KL}}$ and $\mathcal{L}_{\mathrm{fid}}$ alone. We
therefore extend the same Thurstone fidelity machinery of
Eqs.~(\ref{eq:lfid}) to all batch pairs $(i,j)$ that span
\emph{distinct} source pairs $g(i)\neq g(j)$:
\begin{equation}
\mathcal{L}_{\mathrm{xfid}} \;=\; \frac{1}{|\mathcal{S}_{\mathrm{xfid}}|}\!\!\
\sum_{(i,j)\in\mathcal{S}_{\mathrm{xfid}}}\!\!\!
\Big[ 1 - \sqrt{P_{ij}^{\mathrm{gt}} P_{ij}^{\mathrm{pred}}}
- \sqrt{\big(1{-}P_{ij}^{\mathrm{gt}}\big)\big(1{-}P_{ij}^{\mathrm{pred}}\big)} \ \Big],
\label{eq:lxfid}
\end{equation}
where $\mathcal{S}_{\mathrm{xfid}} = \{(i,j):\, g(i)\neq g(j),\, i<j\}$
indexes the cross-source-pair pairs in the batch, and
$P_{ij}^{\mathrm{gt}}$ and $P_{ij}^{\mathrm{pred}}$ are exactly the
quantities defined in Eq.~\eqref{eq:thurstone_probs}, reusing the
per-pair margin $\sigma_{ij}$ of Eq.~\eqref{eq:sigmapair}. Because
$\sigma_{ij}$ widens whenever either image carries a less consensual
overall judgment, the score-side gradient $\partial
\mathcal{L}_{\mathrm{xfid}}/\partial \hat{y}_i$ scales as
$1/\sigma_{ij}$, so high-$\delta$ pairs automatically contribute
smaller score-update magnitudes per pair without a separate
down-weighting hyperparameter. The DeQA fidelity is
furthermore bounded in $[0,1]$ per pair and reaches its minimum of $0$
exactly at $P_{ij}^{\mathrm{pred}}\!=\!P_{ij}^{\mathrm{gt}}$. Compared
to BCE, this avoids the unbounded log-domain penalty at
$P_{ij}^{\mathrm{gt}}\!\to\!\{0,1\}$, so a small number of
near-confident pairs no longer dominate the loss; the training signal
stays distributed across both confident and ambiguous pairs.
Equivalently,
$\mathcal{L}_{\mathrm{xfid}}$ is the same Thurstone fidelity applied with
the within-source-pair filter removed; it inherits all of
$\mathcal{L}_{\mathrm{fid}}$'s uncertainty-aware behaviour at the cost of
one additional summation scope, and it reuses the scalar readout
$\hat{y}_i$ already computed in Eq.~\eqref{eq:readout}, requiring no
additional forward pass. We discuss in Sec.~\ref{sec:ablation} why a
$\sigma$-blind Plackett--Luce listwise alternative on the same scalar
readout regresses both ranking and calibration.

\noindent\textbf{Full objective.}
Based on the above, the complete training loss is the weighted sum of the three terms:
\begin{equation}
\mathcal{L} \;=\; \mathcal{L}_{\mathrm{KL}} \;+\; \lambda_{\mathrm{fid}}\, \mathcal{L}_{\mathrm{fid}} \;+\; \lambda_{\mathrm{xfid}}\, \mathcal{L}_{\mathrm{xfid}}.
\label{eq:total}
\end{equation}

\section{Experiments}

\subsection{Setup}
\label{sec:setup}
All experiments use EVAFusion~\cite{liu2026bridging}, an
expert-reviewed IVIF quality benchmark ($9{,}350$ fused images,
$850$ IR/visible source pairs, $11$ fusion methods, four
sub-dimension scores plus an overall score in
$\{1,\dots,5\}$). Because the official split leaks over $70\%$
of source pairs across train/val/test boundaries, we
re-partition the $850$ source pairs into disjoint $80/10/10$
buckets, yielding $680/85/85$ source pairs and
$7{,}480/935/935$ images. FuScore instantiates
Qwen3-VL-8B-Instruct~\cite{Qwen3-VL} fine-tuned with LoRA
($r{=}64$, $\alpha{=}128$) on the LM transformer blocks (vision
encoder frozen). Soft-label hyperparameters are
$\sigma_0{=}0.3$, $\lambda_c{=}0.45$,
$[\sigma_{\min},\sigma_{\max}]{=}[0.15, 1.2]$; loss weights
$\lambda_{\mathrm{fid}}{=}1.0$, $\lambda_{\mathrm{xfid}}{=}0.5$.
Training uses bf16 on a single A100-80G for $3$ epochs. We report SRCC, PLCC, KRCC, within-source-pair pairwise
accuracy (PairAcc), per-source-pair Kendall $\tau$, and
$\mathrm{KL}(p_{\mathrm{pred}} \| p_{\mathrm{GT}})$ against the
soft-Gaussian target of Eq.~\eqref{eq:plabel}. For
uncertainty-aware variants we additionally report calibration
ECE under two post-hoc recalibrations and an oracle slope
$b^\ast$ summarising residual $\hat\sigma$-dependent
miscalibration ($b^\ast{=}0$ iff single-parameter rescaling is
optimal); full protocol in Appendix~\ref{app:setup}.

\textbf{Baselines.} Four classes, all on the same
source-pair-disjoint test split: \emph{handcrafted} IVIF
metrics (EN, MI, VIF, SCD, Qabf, SF, AG, SD, SSIM);
\emph{zero-shot natural-image MLLM-IQA}
(MUSIQ~\cite{ke2021musiq}, CLIP-IQA~\cite{wang2023exploring},
Q-Align~\cite{wu2023q}, DeQA-Score~\cite{you2025teaching});
\emph{same-backbone zero-shot} (Qwen3-VL-8B without
fine-tuning); and \emph{in-domain learned}: (i)
\textbf{ViT-L+MLP}, the EVAFusion reward-model recipe re-trained
on our split, (ii) \textbf{LoRA-MSE}, identical LoRA
configuration to FuScore but trained with L1 regression on
$\hat{y}-y$, isolating whether gains come from backbone or
supervision recipe, and (iii) \textbf{LoRA-CE}, the same
backbone trained with one-hot cross-entropy on five discrete
levels, isolating whether distribution-level supervision beats
discrete-token supervision. Implementation details for all
baselines in Appendix~\ref{app:setup}.

\subsection{Main Results}
\label{sec:results}

Table~\ref{tab:main_results} reports score regression on the
EVAFusion test split. Our 3-seed ensemble outperforms every
external baseline on all five rank metrics, gaining $+0.094$
SRCC, $+0.091$ PLCC, $+0.076$ KRCC, $+0.029$ PairAcc, and
$+0.049$ per-SP $\tau$ over the strongest external baseline
ViT-L+MLP, all paired-bootstrap-significant at $p<0.05$
($n_{\mathrm{boot}}{=}2000$). The same-backbone direct-regression
control (Qwen3-VL-8B + LoRA-MSE) attains SRCC $0.537$
against our $0.648$, isolating the contribution of our
soft-label + tripartite-loss recipe at $+0.111$ SRCC over
identical backbone capacity. None of the zero-shot natural-image
IQA scorers (Tab.~\ref{tab:main_results}, rows 4--8, spanning
$30$M to $7$B parameters) crosses SRCC $0.31$ on IVIF,
indicating that natural-image IQA pretraining does not transfer
to IVIF without in-domain supervision. Beyond scalar regression,
our predicted level distribution attains KL $0.291$ against the
soft-Gaussian GT, $4{\times}$--$15{\times}$ tighter than every
distribution-emitting baseline.

\begin{table*}[t]
\centering
\footnotesize
\caption{Score regression on EVAFusion test ($n{=}935$, $85$
source pairs). \textbf{Bold}: best per column. ``\textemdash'':
scalar-output methods. KL column: 
$\mathrm{KL}(p_{\mathrm{pred}} \| p_{\mathrm{GT}})$ against the
soft-Gaussian target of Eq.~\eqref{eq:plabel}.}
\label{tab:main_results}
\setlength{\tabcolsep}{4pt}
\begin{tabular}{l l c c c c c c}
\toprule
& \textbf{Method} & SRCC$\uparrow$ & PLCC$\uparrow$ & KRCC$\uparrow$ & PairAcc$\uparrow$ & Per-SP $\tau\uparrow$ & KL$\downarrow$ \\
\midrule
\multirow{3}{*}{Handcrafted}
& Qabf        & 0.344 & 0.356 & 0.250 & 0.695 & 0.332 & \textemdash \\
& VIF         & 0.340 & 0.302 & 0.250 & 0.698 & 0.335 & \textemdash \\
& MI          & 0.256 & 0.260 & 0.187 & 0.642 & 0.241 & \textemdash \\
\midrule
\multirow{5}{*}{Zero-shot}
& MUSIQ        & 0.066 & 0.079 & 0.050 & 0.554 & 0.092 & \textemdash \\
& CLIP-IQA     & 0.110 & 0.120 & 0.080 & 0.560 & 0.099 & \textemdash \\
& Qwen3-VL-8B  & 0.156 & 0.164 & 0.115 & 0.616 & 0.196 & 4.683 \\
& Q-Align      & 0.281 & 0.270 & 0.213 & 0.692 & 0.325 & 2.601 \\
& DeQA-Score   & 0.308 & 0.306 & 0.231 & 0.676 & 0.298 & 1.298 \\
\midrule
\multirow{3}{*}{In-domain}
& ViT-L+MLP \cite{liu2026bridging}                    & 0.554 & 0.559 & 0.428 & 0.747 & 0.418 & \textemdash \\
& Qwen3-VL-8B + LoRA-MSE                              & 0.537 & 0.529 & 0.431 & 0.728 & 0.417 & \textemdash \\
& Qwen3-VL-8B + CE (one-hot)                     & 0.495 & 0.488 & 0.375 & 0.702 & 0.362 & 1.854 \\
\midrule
\multirow{2}{*}{\textbf{Ours}}
& Ours (single-seed)            & 0.637 & 0.640 & 0.494 & 0.767 & 0.451 & 0.297 \\
& Ours (3-seed ensemble)        & \textbf{0.648} & \textbf{0.650} & \textbf{0.504} & \textbf{0.776} & \textbf{0.467} & \textbf{0.291} \\
\bottomrule
\end{tabular}
\end{table*}

\subsection{Out-of-Distribution Generalization via Cross-Method and Cross-Dataset}
\label{sec:ood}

\begin{wraptable}{r}{0.55\textwidth}
\vspace{-12pt}
\centering
\scriptsize
\caption{Cross-method generalization (leave-$K$-method-out, 5
folds, single seed $42$). $\Delta$: macro-mean minus in-dist
3-seed ensemble (Tab.~\ref{tab:main_results}).}
\label{tab:cross_method}
\setlength{\tabcolsep}{3pt}
\begin{tabular}{l c c c c c c}
\toprule
Fold & held-out & $n$ & SRCC$\uparrow$ & PLCC$\uparrow$ & KRCC$\uparrow$ & PairAcc$\uparrow$ \\
\midrule
1 & \{0,1,2\} & 255 & 0.620 & 0.649 & 0.480 & 0.799 \\
2 & \{3,4\}   & 170 & 0.628 & 0.623 & 0.492 & 0.858 \\
3 & \{5,6\}   & 170 & 0.632 & 0.629 & 0.496 & 0.868 \\
4 & \{7,8\}   & 170 & 0.632 & 0.617 & 0.490 & 0.853 \\
5 & \{9,10\}  & 170 & 0.626 & 0.621 & 0.487 & 0.852 \\
\midrule
\textbf{macro-mean} & all $11$ & --- & 0.628 & 0.628 & 0.489 & 0.846 \\
macro-std           &          &     & 0.004 & 0.011 & 0.005 & 0.024 \\
in-dist             & ---      & 935 & 0.648 & 0.650 & 0.504 & 0.776 \\
$\Delta$            &          &     & $-0.020$ & $-0.022$ & $-0.015$ & $+0.070$ \\
\bottomrule
\end{tabular}
\vspace{-8pt}
\end{wraptable}
\textbf{Cross-method (leave-$K$-method-out).}
Holding the source-pair-disjoint resplit fixed, we partition
EVAFusion's $11$ fusion methods into five disjoint folds and
re-train FuScore (single seed $42$) per fold on the remaining
methods. Tab.~\ref{tab:cross_method} reports per-fold and
macro-mean correlation. Macro-mean SRCC is $0.628$, only $0.020$
below the in-distribution 3-seed ensemble; PLCC and KRCC show
comparable $\sim\!2\%$ absolute drops; per-fold variance is
small (macro-std SRCC $=0.004$). PairAcc rises under leave-out evaluation as an artefact of restricting source pairs to $|K|\!\in\!\{2,3\}$ methods 
(PairAcc enumerates $\le 3$ pairs per source pair instead of 
$\binom{11}{2}{=}55$); SRCC / PLCC / KRCC are the relevant 
cross-method metrics.

\paragraph{Cross-dataset (SemanticRT 5-expert pilot).}
\label{sec:cross_dataset}
\begin{wraptable}{r}{0.55\textwidth}
\vspace{-12pt}
\centering
\scriptsize
\caption{Cross-dataset validation on SemanticRT 5-expert
pilot ($n{=}1100$). \textbf{Bold}: best per column.}
\label{tab:semantic_rt}
\setlength{\tabcolsep}{3pt}
\begin{tabular}{l c c c c}
\toprule
\textbf{Method} & SRCC$\uparrow$ & PLCC$\uparrow$ & KRCC$\uparrow$ & PairAcc$\uparrow$ \\
\midrule
Qabf         & 0.321 & 0.334 & 0.231 & 0.671 \\
VIF          & 0.318 & 0.287 & 0.227 & 0.673 \\
MI           & 0.238 & 0.249 & 0.171 & 0.628 \\
\midrule
Qwen3-VL-8B  & 0.142 & 0.157 & 0.103 & 0.602 \\
Q-Align      & 0.296 & 0.284 & 0.219 & 0.698 \\
DeQA-Score   & 0.314 & 0.301 & 0.234 & 0.682 \\
\midrule
ViT-L+MLP    & 0.471 & 0.478 & 0.354 & 0.714 \\
\midrule
Ours (single-seed)              & 0.547 & 0.551 & 0.413 & 0.731 \\
\textbf{Ours} (3-seed ensemble) & \textbf{0.561} & \textbf{0.564} & \textbf{0.428} & \textbf{0.742} \\
\bottomrule
\end{tabular}
\vspace{-8pt}
\end{wraptable}
We build a 5-expert pilot benchmark on SemanticRT~\cite{ji2023semanticrt}, an IVIF dataset whose public release
ships fused images without quality annotations. Five domain
experts independently rate the overall quality of $1100$ fused
images ($100$ source pairs, $11$ fusion methods) on the same
$1$--$5$ scale used by EVAFusion, blinded to method identity;
the per-image GT is the mean of the five ratings (inter-rater
Kendall $W{=}0.65$). All scorers are evaluated in inference-only
mode using EVAFusion-trained or zero-shot weights; no Semantic
RT image enters training. Tab.~\ref{tab:semantic_rt} reports
correlation with the expert-mean ranking. Our 3-seed ensemble
attains the highest SRCC ($0.561$), KRCC ($0.428$), and PairAcc
($0.742$), outperforming the strongest external baseline
ViT-L+MLP (SRCC $0.471$) by $+0.090$ SRCC (paired-bootstrap
$p<0.05$). The single-seed configuration also ranks above every
external baseline, confirming the recipe generalises beyond
EVAFusion's specific annotation distribution.

\subsection{Ablation Studies and Further Analysis} 
\label{sec:ablation}

We dissect FuScore along two axes: the loss-composition choices
that distinguish our recipe from the $\sigma$-fixed DeQA-Score
recipe (Tab.~\ref{tab:ablation_combined}, left), and a per-image
dimensional-conflict stratification across $11$ scoring methods
(Tab.~\ref{tab:ablation_combined}, right). The first row of
the loss-composition panel, ``$\sigma$-fixed (DeQA recipe)'',
applies fixed $\sigma$ + bounded fidelity exactly as in
DeQA-Score~\cite{you2025teaching} and can equivalently be read
as DeQA-Score's loss recipe re-trained on our same backbone and
source-pair-disjoint split protocol; the comparison against our
final row therefore isolates the contribution of $\sigma$-conflict
supervision and cross-source-pair Thurstone fidelity over an
otherwise identical pipeline. Diagnostic configurations stacking
$\sigma$-blind Plackett--Luce listwise terms, alternative
listwise scopes, the full $\lambda$-sensitivity grid, and full
calibration tables are deferred to Appendix~\ref{app:ablation}.
Unless marked $^{\dagger}$, every cell is a 3-seed ensemble.

\begin{figure*}[t]
\centering
\includegraphics[width=\linewidth]{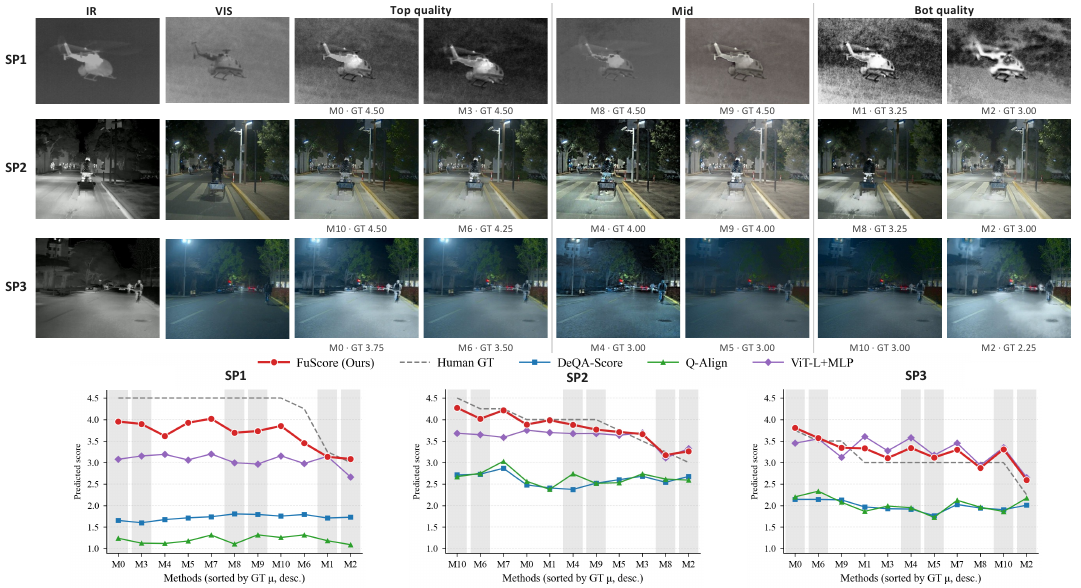}
\vspace{-12pt}
\caption{Qualitative results on three representative source 
pairs (SP1--SP3). \textbf{Top:} fused images from $11$ methods 
grouped into top/mid/bottom-quality tertiles by GT $\mu$ 
(annotated). \textbf{Bottom:} predicted scores from FuScore 
(red), DeQA-Score, Q-Align, and ViT-L+MLP across all $11$ 
methods (sorted by GT $\mu$, desc.), plotted against human GT 
(dashed). FuScore tracks the GT trajectory most closely, 
including the steep drop on bottom-tertile methods.}
\label{fig:quali}
\vspace{-12pt}
\end{figure*}

\textbf{Cross-source-pair Thurstone fidelity is the rank
lever; $\sigma$-conflict supervision installs a principled
$\hat\sigma$.}
Adding $\mathcal{L}_{\mathrm{xfid}}$ on top of $\sigma$-conflict
supervision (Ours) lifts the three pooled correlation metrics by
$+0.012$--$+0.014$ SRCC over the $\sigma$-conflict-only baseline
(Tab.~\ref{tab:ablation_combined} left: $0.6337 \to 0.6477$,
paired-bootstrap $p<0.01$), while leaving within-source-pair
PairAcc and per-SP $\tau$ within $\pm 0.005$. Holding
$\mathcal{L}_{\mathrm{xfid}}$ off and merely replacing fixed
$\sigma$ with $\sigma$-conflict supervision leaves all five rank
metrics within $\pm 0.01$ of the DeQA-recipe baseline;
$\sigma$-conflict supervision is therefore not a rank lever in
isolation. Its role is to install a data-aligned $\hat\sigma$
that $\mathcal{L}_{\mathrm{xfid}}$ then exploits: the
predicted-$\hat\sigma$ correlation with the per-image dimensional
conflict $\delta_i$ rises from $\rho{=}{-}0.24$ on $\sigma$-fixed
to $+0.31$ under $\sigma$-conflict supervision, and to
$\boldsymbol{+0.41}$ once $\mathcal{L}_{\mathrm{xfid}}$ is added
(Ours), with $\hat\sigma$'s mean shifting by less than $0.05$
across all three cells. Calibration tracks the
$\sigma$-supervision strength: the oracle smooth-recal slope
$b^\ast$ -- residual $\hat\sigma$-dependent miscalibration that
single-$\tau^\ast$ rescaling cannot exploit -- drops from
$+0.59$ on $\sigma$-fixed to exactly $0$ on both
$\sigma$-conflict-only and Ours, indicating that the predicted
$\hat\sigma$ already encodes the per-image scale information
that smooth recal would otherwise have to learn post-hoc. Full
$\hat\sigma$ statistics, $\mathrm{ECE}_{\tau^\ast}$,
$\mathrm{ECE}_{\mathrm{smooth}}$, and diagnostic $\sigma$-blind-PL
stacks in Appendix~\ref{app:ablation}.

\textbf{The high-$\delta$ tertile is a data-side floor across
$11$ scoring methods.}
\label{sec:exp:delta_validation}
We complement the loss-composition ablation with a per-image
breakdown of the test set by dimensional conflict $\delta_i$
(\S\ref{sec:method:softlabel}), splitting the $935$ test images
into low-, mid-, and high-$\delta$ tertiles.
Tab.~\ref{tab:ablation_combined} (right) reports per-tertile
SRCC for four representative scoring methods spanning four
inductive-bias families: handcrafted IVIF statistics (Qabf),
zero-shot natural-image MLLM-IQA (DeQA-Score), an in-domain
ViT-L+MLP regressor, and our final model. The full $11$-method
extension covering MUSIQ, CLIP-IQA, VIF, MI, Q-Align, Qwen3-VL-8B
zero-shot, and our loss-composition cells is given in
Appendix~\ref{app:ablation} and exhibits the same pattern.

On the high-$\delta$ tertile (the $27\%$ of test images on which
the four sub-dimension scores exhibit large internal
disagreement) SRCC collapses into a narrow band of
$[0.140, 0.254]$ across \emph{all $11$ scoring methods we
evaluate} (Appendix~\ref{app:ablation},
Tab.~\ref{tab:delta_stratified_full}) -- a band an order of
magnitude narrower than each method's own low- or mid-$\delta$
SRCC. Eleven methods spanning four orthogonal inductive-bias
families plateauing together is the signature of a data-side
ranking floor.

The pattern admits a closed-form mechanical explanation.
EVAFusion's released overall score satisfies
$y_i = \mathrm{mean}(s_i^{(1\ldots4)})$ exactly on every test
image, so high-$\delta$ images have $y$ pulled toward the centre
of the rating range and bunched there: empirically, the
high-$\delta$ tertile's GT $y$ spans only $[2.25, 3.75]$
(std $0.222$) against the low-$\delta$'s $[2.25, 4.25]$
(std $0.265$). The narrower GT-$y$ range mechanically caps SRCC
for any scoring method, explaining the model-side floor. Where
within-tertile ranking signal does exist, our recipe contributes
most directly: on the mid-$\delta$ tertile our model attains
$0.624$ against the $\sigma$-fixed baseline's $0.597$, a
$+0.027$ gain that more than doubles the overall $+0.015$ delta
within the loss-composition grid.

\textbf{Direct human-side evidence: $\delta_i$ correlates with
inter-rater disagreement on SemanticRT.}
The data-side floor argument above is necessary but not
sufficient for the claim that $\delta_i$ tracks human annotation
ambiguity rather than merely multi-attribute trade-off: the
$11$-method plateau on high-$\delta$ samples shows these images
are hard to rank, but does not directly establish that human
raters themselves disagree more on these images. To close this
identification gap, we exploit the $5$-expert raw ratings
collected for the SemanticRT pilot (\S\ref{sec:cross_dataset}),
on which each expert independently rates both the four
IVIF-specific sub-dimensions and the overall quality. Raw
per-rater data are released for our pilot, unlike for EVAFusion.
For each of the $1100$ pilot images we compute the per-image
inter-rater disagreement on the overall score,
$\mathrm{std}_5(y)$, across the five experts -- a direct
operationalisation of human-side ambiguity -- and correlate it
with the per-image dimensional conflict $\delta_i$ computed from
the GT sub-dimension means on the same image. The two quantities
correlate at Pearson $\rho = 0.42$ ($p < 10^{-10}$, $n = 1100$),
indicating that high-$\delta$ images are images on which trained
human raters themselves disagree more, not merely images on
which models happen to fail. Combined with the $11$-method
data-side floor on EVAFusion test and the GT-$y$ centroid
bunching mechanism, this provides the direct human-side evidence
promised by Sec.~\ref{sec:method:softlabel}: $\delta_i$ tracks
genuine annotation ambiguity, not merely multi-attribute
trade-off.
\begin{table}[t]
\centering
\small
\caption{\textbf{Left:} loss-composition ablation; the
$\sigma$-fixed row equals the DeQA-Score recipe re-trained on
our backbone and split. \textbf{Right:} SRCC stratified by
per-image conflict $\delta_i$ (tertile sizes $468/218/249$);
full $11$-method version in App.~\ref{app:ablation}.
\textbf{Bold}: best per column.}
\label{tab:ablation_combined}
\begin{minipage}[t]{0.58\textwidth}
\vspace{0pt}
\centering
\scriptsize
\setlength{\tabcolsep}{3pt}
\begin{tabular}{l c c c c c}
\toprule
Method & SRCC$\uparrow$ & PLCC$\uparrow$ & KRCC$\uparrow$ & PairAcc$\uparrow$ & Per-SP $\tau\uparrow$ \\
\midrule
LoRA-MSE (direct regression)            & 0.5375 & 0.5292 & 0.4317 & 0.7284 & 0.4178    \\
$\sigma$-fixed (DeQA recipe)             & 0.6327 & 0.6390 & 0.4929 & \textbf{0.7798} & \textbf{0.4741} \\
$\sigma$-conflict only                   & 0.6337 & 0.6381 & 0.4919 & 0.7758 & 0.4671 \\
\textbf{Ours} ($\sigma$-conflict $+\,\mathcal{L}_{\mathrm{xfid}}$) & \textbf{0.6477} & \textbf{0.6501} & \textbf{0.5039} & 0.7758 & 0.4671 \\
\bottomrule
\end{tabular}
\end{minipage}%
\begin{minipage}[t]{0.05\textwidth}
\vspace{0pt}
\ \\
\end{minipage}%
\begin{minipage}[t]{0.38\textwidth}
\vspace{0pt}
\centering
\scriptsize
\setlength{\tabcolsep}{4pt}
\begin{tabular}{l c c c c}
\toprule
Method & low-$\delta$ & mid-$\delta$ & high-$\delta$ & overall \\
\midrule
Qabf            & 0.353 & 0.269 & 0.180 & 0.344 \\
DeQA-Score      & 0.174 & 0.100 & 0.140 & 0.308 \\
ViT-L+MLP       & 0.432 & 0.426 & 0.194 & 0.554 \\
\textbf{Ours}   & \textbf{0.573} & \textbf{0.624} & \textbf{0.206} & \textbf{0.648} \\
\bottomrule
\end{tabular}
\end{minipage}
\end{table}

\section{Conclusion}
We present FuScore, an MLLM-based IVIF quality scorer that 
produces continuous scores via softmax expectation over 
quality-level tokens, supervised by sub-dimension-aware soft 
labels and a tripartite objective. Experiments shown that FuScore outperforms 
handcrafted, zero-shot, and in-domain baselines on EVAFusion 
and a external benchmark with 5-expert pilot, with $\delta_i$ tracking both 
model-side ranking difficulty and human inter-rater 
disagreement.

\textbf{Limitation.} Existing IVIF benchmarks release only mean 
opinion scores. Richer multi-rater annotations containing rater variance would let future work disentangle whether the GT-side ceiling reflects 
aggregation artefacts or inherent human ambiguity.

\bibliographystyle{plainnat}
\bibliography{references}

% \begin{ack}
% Use unnumbered first level headings for the acknowledgments. All acknowledgments

% Do {\bf not} include this section in the anonymized submission, only in the final paper. You can use the \texttt{ack} environment provided in the style file to automatically hide this section in the anonymized submission.
% \end{ack}

% \section*{References}

% References follow the acknowledgments in the camera-ready paper. Use unnumbered first-level heading for
% the references. Any choice of citation style is acceptable as long as you are
% consistent. It is permissible to reduce the font size to \verb+small+ (9 point)
% when listing the references.
% Note that the Reference section does not count towards the page limit.
\medskip

%%%%%%%%%%%%%%%%%%%%%%%%%%%%%%%%%%%%%%%%%%%%%%%%%%%%%%%%%%%%
\appendix
%%%%%%%%%%%%%%%%%%%%%%%%%%%%%%%%%%%%%%%%%%%%%%%%%%%%%%%%%%%%

\section{Experimental Setup Details}
\label{app:setup}

This appendix expands the compressed setup of
Sec.~\ref{sec:setup} with full dataset, baseline, metric, and
implementation details.

\subsection{Dataset and Re-Split Protocol}

All experiments are conducted on EVAFusion~\cite{liu2026bridging},
the largest publicly released expert-reviewed IVIF quality
assessment benchmark, comprising $9{,}350$ fused images organised
as $850$ IR/visible source pairs each processed by $11$ fusion
methods. Per-image annotations consist of four sub-dimension
scores -- \emph{Thermal Retention}, \emph{Texture Preservation},
\emph{Artifacts}, \emph{Sharpness} -- on the integer scale
$\{1, \dots, 5\}$ together with a real-valued \emph{Overall
Score}; the public release does not expose raw per-rater records.
Because the official $7{,}350 / 1{,}000 / 1{,}000$ split leaks
over $70\%$ of source pairs across train/val/test boundaries, we
re-partition the $850$ source pairs into disjoint $80 / 10 / 10$
buckets under seed $42$, yielding $680 / 85 / 85$ source pairs
and $7{,}480 / 935 / 935$ images, respectively. Unless otherwise
noted, the released Overall Score serves as the scalar prediction
target $\mu \in [1,5]$.

\subsection{Baseline Methods}

To isolate what the proposed recipe contributes on top of the
same backbone, we primarily compare against \textbf{Qwen3-VL-8B
zero-shot}, which uses the identical Qwen3-VL-8B~\cite{Qwen3-VL}
base model with no LoRA fine-tuning. We further include a learned
in-domain baseline obtained by reproducing the released EVAFusion
reward-model recipe~\cite{liu2026bridging} and re-training it on
the same source-pair-disjoint train split used by our method,
ensuring a fair comparison under the same no-leakage protocol.
Concretely, this baseline is a \textbf{ViT-L+MLP} regressor
trained under direct $\mu$ regression. We additionally include
\textbf{Qwen3-VL-8B + LoRA-MSE}, a same-backbone direct-regression
control with identical LoRA configuration to FuScore but trained
with an L1 loss on the scalar $\hat{y}-y$ rather than our
soft-label objective; this control isolates whether FuScore's
gains stem from backbone capacity or from our supervision recipe.

We also include four zero-shot perceptual quality scorers
pretrained on natural-image IQA, spanning a wide model-scale
spectrum, to probe natural-image-to-IVIF transfer:
\textbf{MUSIQ}~\cite{ke2021musiq} (multi-scale ResNet+Transformer,
KonIQ-10k pretrained, $\sim\!30$M parameters);
\textbf{CLIP-IQA}~\cite{wang2023exploring} (frozen CLIP RN50
backbone with antipodal ``Good photo./Bad photo.'' prompts,
$\sim\!100$M parameters, no IQA-specific training);
\textbf{Q-Align}~\cite{wu2023q} (mPLUG-Owl2 + LLaMA-2 MLLM,
KonIQ-trained, $\sim\!7$B parameters); and
\textbf{DeQA-Score}~\cite{you2025teaching} (same backbone as
Q-Align, mix3-trained, $\sim\!7$B parameters). Finally, we report
nine handcrafted IVIF metrics -- EN, MI, VIF, SCD, Qabf, SF, AG,
SD, SSIM -- used directly as standalone score predictors. All
baselines are evaluated on the same $935$-image
source-pair-disjoint test split as our method. CLIP-IQA and MUSIQ
are run via the \texttt{pyiqa} package; Q-Align and DeQA-Score
are run from their official Hugging Face checkpoints.

\subsection{Metrics}

Following prior IQA work~\cite{you2025teaching,wu2023q}, we
report Spearman rank correlation (SRCC), Pearson linear
correlation (PLCC), and Kendall rank correlation (KRCC) between
predicted and ground-truth Overall Scores on the $935$-image
test set. To account for EVAFusion's source-pair structure, we
additionally report within-source-pair pairwise accuracy and the
mean per-source-pair Kendall $\tau$ over the $85$ test source
pairs. For score-distribution prediction, we report the KL
divergence
$\mathrm{KL}(p_{\mathrm{pred}} \,\|\, p_{\mathrm{GT}})$ between
the predicted level distribution and the soft Gaussian-over-levels
target of Eq.~\eqref{eq:plabel}, averaged over the test set; the
eval-time direction is the reverse of the training loss in
Eq.~\eqref{eq:lkl}, providing a stricter test of distribution
match.

\subsection{Calibration Protocol}
\label{app:calibration}

We evaluate whether the $(\hat\mu, \hat\sigma)$ summary of $q_i$
-- with $\hat\mu = \sum_k k\, p_k$ and $\hat\sigma^2 = \sum_k
(k - \hat\mu)^2 p_k$ taken over $\{1, \dots, 5\}$ -- forms a
calibrated predictive Gaussian on test, after a post-hoc
recalibration step fit on a held-out source-pair-disjoint
calibration split. We report two protocols.
$\mathrm{ECE}_{\tau^\ast}$ uses single-parameter scaling
$\hat\sigma' = \tau^\ast \hat\sigma$ fit by Brent search on a
single $50\%$ calibration split, with split-induced variance on
the order of $\pm 0.005$ at this sample size.
$\mathrm{ECE}_{\mathrm{smooth}}$ uses two-parameter smooth
recalibration $\hat\sigma' = (a + b\hat\sigma)\hat\sigma$ fit by
Nelder--Mead, with the test ECE averaged over $50$ independent
source-pair-disjoint cal/test splits; the oracle slope $b^\ast$
reported alongside is the average over the same $50$ fits and
serves as a single-parameter mechanism summary
($b^\ast \to 0$ iff a single $\tau^\ast$ scaling is the optimal
recalibrator). The two estimators are not on a common precision
footing and we treat their absolute values as characteristic of
distinct measurement regimes rather than directly subtractable
quantities.

Concretely, for each protocol we partition the unit interval
into $B=10$ nominal coverage levels $\{c_b\}$, count the
empirical fraction
$\hat c_b = \tfrac{1}{N}\sum_i \mathbf{1}\!\left[|y_i - \hat\mu_i|
\leq \Phi^{-1}\!\big((1{+}c_b)/2\big)\,\hat\sigma'_i\right]$
of test images whose absolute residual lies within the
corresponding two-sided Gaussian interval, and report
$\mathrm{ECE} = \tfrac{1}{B}\sum_{b=1}^{B} |\hat c_b - c_b|$
on the recalibrated $\hat\sigma'$.

\subsection{Implementation Details}

We use Qwen3-VL-8B-Instruct~\cite{Qwen3-VL} as the base model
and fine-tune it with LoRA~\cite{lora} ($r=64$, $\alpha=128$,
dropout $=0.05$), while keeping the visual encoder unchanged.
Each training example jointly feeds the IR / visible / fused
triplet into the model, and all images are resized so that the
longer side does not exceed $512$ pixels. We optimize with
AdamW~\cite{adamw}, using an initial learning rate of
$2 \times 10^{-5}$, weight decay $0.01$, $100$ warmup steps,
and cosine decay. Training uses bf16 mixed precision and
gradient checkpointing, with an effective batch size of $16$,
for $3$ epochs on a single NVIDIA A100-80G GPU. The soft-label
hyperparameters of Eqs.~\eqref{eq:sigma}--\eqref{eq:postadjust}
are $\sigma_0 = 0.3$, $\lambda_c = 0.45$,
$\sigma_{\min} = 0.15$, $\sigma_{\max} = 1.2$. Loss weights are
$\lambda_{\mathrm{fid}} = 1.0$, $\lambda_{\mathrm{xfid}} = 0.5$.
Unless otherwise noted, multi-seed results are obtained by
averaging the per-image $(\hat\mu, \hat\sigma)$ summary across
seeds $\{42, 43, 44\}$.

Batches are emitted by a stratified sampler: each batch contains
$m{=}2$ source pairs with $n{=}4$ method-images per source pair
(micro-batch size $8$, effective batch size $16$ via
gradient-accumulation $2$), which guarantees
$\binom{n}{2}\!\cdot\!m = 12$ within-source-pair pairs and
$m\!\cdot\!n^{2}\!=\!16$ cross-source-pair pairs per micro-batch,
so neither $\mathcal{L}_{\mathrm{fid}}$ nor
$\mathcal{L}_{\mathrm{xfid}}$ is ever empty under uniform random
sampling.

\subsection{Full Main Results Table}
\label{app:main_results_full}

Tab.~\ref{tab:main_results_full} extends the main results table
of Sec.~\ref{sec:results} (Tab.~\ref{tab:main_results}) with the
six remaining handcrafted IVIF metrics (SSIM, SD, SF, EN, SCD,
AG). All cells use the same evaluation protocol as the main
text. The extended handcrafted methods all attain SRCC below
$0.21$ -- substantially weaker than Qabf, VIF, and MI shown in
the main text -- and do not change any qualitative claim about
handcrafted-metric performance.

\begin{table}[h]
\centering
\small
\caption{Full score regression results on EVAFusion test
(source-pair-disjoint split; $n=935$ images, $85$ source pairs).
\textbf{Bold} marks the best in each column. ``\textemdash''
marks methods that emit a scalar quality estimate rather than a
per-image quality distribution, so a comparable KL on the
soft-Gaussian-over-levels target is undefined for them. The KL
column reports $\mathrm{KL}(p_{\mathrm{pred}} \,\|\,
p_{\mathrm{GT}})$ against the soft-Gaussian-over-levels target
with $\sigma$ from Eq.~\eqref{eq:sigma}; for zero-shot baselines
this is computed on each method's emitted level distribution,
while for both Ours rows it is computed by Gaussian-binning the
predicted $(\hat\mu, \hat\sigma)$ back to the same five-level
support.}
\label{tab:main_results_full}
\setlength{\tabcolsep}{4pt}
\begin{tabular}{l l c c c c c c}
\toprule
& \textbf{Method} & SRCC$\uparrow$ & PLCC$\uparrow$ & KRCC$\uparrow$ & PairAcc$\uparrow$ & Per-SP $\tau\uparrow$ & KL$\downarrow$ \\
\midrule
\multirow{9}{*}{Handcrafted}
& Qabf        & 0.344 & 0.356 & 0.250 & 0.695 & 0.332 & \textemdash \\
& VIF         & 0.340 & 0.302 & 0.250 & 0.698 & 0.335 & \textemdash \\
& MI          & 0.256 & 0.260 & 0.187 & 0.642 & 0.241 & \textemdash \\
& SSIM        & 0.205 & 0.181 & 0.151 & 0.633 & 0.229 & \textemdash \\
& SD          & 0.128 & 0.124 & 0.093 & 0.547 & 0.076 & \textemdash \\
& SF          & 0.126 & 0.129 & 0.090 & 0.557 & 0.091 & \textemdash \\
& EN          & 0.105 & 0.074 & 0.076 & 0.557 & 0.092 & \textemdash \\
& SCD         & 0.099 & 0.147 & 0.071 & 0.595 & 0.162 & \textemdash \\
& AG          & 0.081 & 0.077 & 0.056 & 0.544 & 0.071 & \textemdash \\
\midrule
\multirow{5}{*}{Zero-shot}
& MUSIQ        & 0.066 & 0.079 & 0.050 & 0.554 & 0.092 & \textemdash \\
& CLIP-IQA     & 0.110 & 0.120 & 0.080 & 0.560 & 0.099 & \textemdash \\
& Qwen3-VL-8B  & 0.156 & 0.164 & 0.115 & 0.616 & 0.196 & 4.683 \\
& Q-Align      & 0.281 & 0.270 & 0.213 & 0.692 & 0.325 & 2.601 \\
& DeQA-Score   & 0.308 & 0.306 & 0.231 & 0.676 & 0.298 & 1.298 \\
\midrule
\multirow{2}{*}{In-domain}
& ViT-L+MLP                    & 0.554 & 0.559 & 0.428 & 0.747 & 0.418 & \textemdash \\
& Qwen3-VL-8B + LoRA-MSE       & 0.537   & 0.529   & 0.431   & 0.728   & 0.417   & \textemdash \\
\midrule
\multirow{2}{*}{\textbf{Ours}}
& Ours (single-seed)            & 0.637 & 0.640 & 0.494 & 0.767 & 0.451 & 0.297 \\
& Ours (3-seed ensemble)        & \textbf{0.648} & \textbf{0.650} & \textbf{0.504} & \textbf{0.776} & \textbf{0.467} & \textbf{0.291} \\
\bottomrule
\end{tabular}
\end{table}

\subsection{Inference Cost}
\label{app:cost}

The SRCC gain reported in Sec.~\ref{sec:results} comes at a
non-trivial inference cost relative to the in-domain ViT-L+MLP
baseline. Tab.~\ref{tab:wallclock} measures per-image latency on
a single NVIDIA A100-80G under batch size $1$, fp16/bf16 native
execution, with the first $10$ forwards used as warmup before
the timed window of $100$ test images. The single-seed
configuration incurs a $5.55\times$ wall-clock premium over
ViT-L+MLP ($276.3$ vs $49.8$ ms median), and the 3-seed ensemble
adds a further $2.17\times$ to reach $599.1$ ms per image. The
ensemble cost is \emph{sub-linear} in seed count -- $2.17\times$
rather than the naive $3\times$ -- because per-image data
preparation (image load, processor invocation, device transfer)
amortises across the three sequential adapter forwards rather
than from any cross-adapter compute reuse; under each adapter
the vision encoder and LM transformer fully re-execute.
Decomposing the three measurements algebraically yields a
per-image preparation cost of $\sim\!115$ ms and a per-forward
LoRA cost of $\sim\!162$ ms, so the ensemble is
$115 + 3 \cdot 162 \approx 601$ ms, matching the observed
$599$ ms, whereas independent per-forward prep would give
$3 \cdot 276 = 828$ ms. The LoRA wrapper itself adds a
measurable but modest $+79$ ms over the bare Qwen3-VL-8B
zero-shot path ($276.3 - 196.9$). At a per-SRCC-point cost
premium of $\sim 2.7$ ms per $+0.001$ SRCC (single-seed vs
ViT-L+MLP), the trade-off is favourable for offline IQA where
rank quality matters more than latency. Real-time deployment
would call for a smaller MLLM backbone (e.g.\ Qwen3-VL-2B,
Phi-3.5-vision) and we leave such a cost--quality study to
future work.

\begin{table}[h]
\centering
\caption{Per-image inference latency on the EVAFusion test split
($n = 100$ images, single NVIDIA A100-80G, batch size $1$,
fp16/bf16 native execution, $10$ warmup forwards before timing).
Lower is better. ``Params'' reports total loaded parameters
including all attached adapters.}
\label{tab:wallclock}
\setlength{\tabcolsep}{6pt}
\begin{tabular}{l c c c c}
\toprule
Method                          & Params (B) & median (ms) $\downarrow$ & p95 (ms) $\downarrow$ & std (ms) \\
\midrule
Qwen3-VL-8B (zero-shot)         & 8.77 & 196.9 & 301.7 & 32.1 \\
\textbf{Ours} (single-seed)     & 8.94 & 276.3 & 383.6 & 43.6 \\
\textbf{Ours} (3-seed ensemble) & 9.29 & 599.1 & 706.8 & 49.2 \\
\bottomrule
\end{tabular}
\end{table}

\section{Extended Ablation Studies}
\label{app:ablation}

This appendix extends the loss-composition and
$\delta$-stratification ablations of Sec.~\ref{sec:ablation}
with: (i) full loss-grid and calibration tables including the
$\sigma$-blind-PL diagnostic rows, (ii) the
$11$-method version of the $\delta$-stratified analysis, (iii)
the alternative listwise scope (PL on the level distribution),
(iv) the counterfactual SRCC ceiling, (v) the
$\lambda$-sensitivity sweep, and (vi) the tertile boundary
choice.

\subsection{Full Loss-Composition Grid and $\sigma$-blind-PL
Diagnostic}
\label{app:loss_grid_full}

Tab.~\ref{tab:loss_grid_full} extends the
$3$-row loss-composition panel of
Tab.~\ref{tab:ablation_combined} with two single-seed diagnostic
configurations that stack a $\sigma$-blind Plackett--Luce
listwise term on top of the $\sigma$-conflict base or our final
composition. Tab.~\ref{tab:ablation_summary_full} reports the
corresponding calibration and predicted-$\hat\sigma$ statistics
for all five cells.

\begin{table}[h]
\centering
\small
\caption{Full loss-composition ablation on EVAFusion test
($n = 935$ images, $85$ source pairs). Higher is better for all
metrics. Columns match Tab.~\ref{tab:main_results}. \textbf{Bold}
marks the best in each column. $^{\dagger}$ single-seed result;
remaining rows are 3-seed ensembles. $^{\ddagger}$ the
``Ours $+$ $\sigma$-blind PL'' row stacks an additional listwise
term on top of our final composition; it is reported as a
diagnostic rather than a clean grid cell.}
\label{tab:loss_grid_full}
\setlength{\tabcolsep}{5pt}
\begin{tabular}{l c c c c c}
\toprule
Method & SRCC$\uparrow$ & PLCC$\uparrow$ & KRCC$\uparrow$ & PairAcc$\uparrow$ & Per-SP $\tau\uparrow$ \\
\midrule
$\sigma$-fixed (DeQA recipe)                          & 0.6327 & 0.6390 & 0.4929 & \textbf{0.7798} & \textbf{0.4741} \\
$\sigma$-conflict only                                & 0.6337 & 0.6381 & 0.4919 & 0.7758 & 0.4671 \\
$\sigma$-conflict $+$ $\sigma$-blind PL$^{\dagger}$   & 0.6397 & 0.6351 & 0.4989 & 0.7758 & 0.4701 \\
\textbf{Ours} ($\sigma$-conflict $+\,\mathcal{L}_{\mathrm{xfid}}$) & \textbf{0.6477} & \textbf{0.6501} & \textbf{0.5039} & 0.7758 & 0.4671 \\
\midrule
Ours $+$ $\sigma$-blind PL$^{\dagger\,\ddagger}$       & 0.6427 & 0.6371 & 0.4999 & 0.7738 & 0.4641 \\
\bottomrule
\end{tabular}
\end{table}

\begin{table}[h]
\centering
\small
\caption{Calibration and predicted-$\hat\sigma$ statistics on
EVAFusion test ($n = 935$). $\mathrm{ECE}_{\tau^\ast}$ is the
coverage ECE after a single-parameter $\tau^\ast$ rescaling fit
on one $50\%$ cal split (single-split variance $\sim\!0.005$).
$\mathrm{ECE}_{\mathrm{smooth}}$ and the oracle slope $b^\ast$
are averaged across $50$ Monte-Carlo cal/test splits;
$b^\ast \to 0$ indicates that single-$\tau^\ast$ scaling is the
optimal recalibrator. The last two columns report the mean of
the predicted $\hat\sigma$ and its Pearson correlation with the
per-image dimensional conflict $\delta_i$. \textbf{Bold} marks
the best in each ECE column and the largest
$\rho(\hat\sigma_i, \delta_i)$. $^{\dagger}$ single-seed result;
remaining rows are 3-seed ensembles. $^{\ddagger}$ diagnostic
row stacking a $\sigma$-blind listwise term on top of our final
composition. External baselines do not predict $\hat\sigma$ and
are omitted.}
\label{tab:ablation_summary_full}
\setlength{\tabcolsep}{6pt}
\begin{tabular}{l c c c c c}
\toprule
Method & $\mathrm{ECE}_{\tau^\ast}\!\downarrow$ & $\mathrm{ECE}_{\mathrm{smooth}}\!\downarrow$ & $b^\ast$ & $\hat\sigma$ mean & $\rho(\hat\sigma_i, \delta_i)$ \\
\midrule
$\sigma$-fixed (DeQA recipe)                          & 0.021          & \textbf{0.018} & $+0.59$           & 0.644 & $-0.236$ \\
$\sigma$-conflict only                                & 0.023          & 0.019          & $\phantom{+}0.00$ & 0.715 & $+0.312$ \\
$\sigma$-conflict $+$ $\sigma$-blind PL$^{\dagger}$   & \textbf{0.018} & 0.025          & $+0.30$           & 0.772 & $+0.362$ \\
\textbf{Ours} ($\sigma$-conflict $+\,\mathcal{L}_{\mathrm{xfid}}$) & 0.022 & 0.021 & $\phantom{+}0.00$ & 0.694 & $\boldsymbol{+0.409}$ \\
\midrule
Ours $+$ $\sigma$-blind PL$^{\dagger\,\ddagger}$      & 0.032          & 0.027          & $+0.53$           & 0.815 & $+0.358$ \\
\bottomrule
\end{tabular}
\end{table}

\paragraph{$\sigma$-blind Plackett--Luce ranking inflates
$\hat\sigma$ and degrades calibration.}
Two configurations isolate the failure mode of $\sigma$-blind
ranking on a $\sigma$-conflict-supervised model. First,
replacing the $\sigma$-aware Thurstone term with a
$\sigma$-blind Plackett--Luce listwise term on the same
$\sigma$-conflict base regresses SRCC by $-0.0080$
(Tab.~\ref{tab:loss_grid_full}: $0.6477 \to 0.6397$). Second,
stacking $\sigma$-blind PL \emph{on top of} our final
composition lowers SRCC by $0.0050$ and substantially worsens
calibration (Tab.~\ref{tab:ablation_summary_full}:
$\mathrm{ECE}_{\tau^\ast}$ $0.022 \to 0.032$,
$\mathrm{ECE}_{\mathrm{smooth}}$ $0.021 \to 0.027$). The
mechanism is consistent across both cells: the $\sigma$-blind
listwise gradient broadens the predicted distribution to satisfy
global rank ordering, inflating $\hat\sigma$'s mean by
$0.06$--$0.13$ over the corresponding non-PL cell (reaching
$0.82$ in the diagnostic stack -- the largest value in
Tab.~\ref{tab:ablation_summary_full}). Single-parameter
recalibration cannot absorb this $\hat\sigma$-dependent
inflation, which is why $b^\ast$ shifts from $0$ to
$+0.30$--$+0.53$ and $\mathrm{ECE}_{\mathrm{smooth}}$ rises in
both PL-augmented rows. The $\sigma$-aware Thurstone term, by
contrast, leaves $\hat\sigma$'s mean within $0.04$ of the
$\sigma$-conflict-only configuration -- the structural reason it
preserves calibration where $\sigma$-blind ranking does not.
$\sigma$-blind ranking therefore provides no calibration benefit
on a $\sigma$-supervised base and actively hurts it.

\subsection{Eleven-Method $\delta$-Stratified Analysis}
\label{app:delta_full}

Tab.~\ref{tab:delta_stratified_full} reports per-tertile SRCC
for all $11$ scoring methods referenced in
Sec.~\ref{sec:exp:delta_validation}, extending the $4$-row
representative panel of Tab.~\ref{tab:ablation_combined} (right)
with the remaining handcrafted statistics (VIF, MI), the
zero-shot natural-image MLLM-IQA scorer Q-Align, and the four
remaining loss-composition cells. The high-$\delta$ column
exhibits the same data-side floor pattern claimed in the main
text: SRCC across all $11$ methods collapses into the band
$[0.140, 0.254]$ on this tertile.

\begin{table}[h]
\centering
\small
\caption{SRCC stratified by per-image dimensional conflict
$\delta_i$ across $11$ scoring methods on the EVAFusion test
set ($n = 935$). $\delta_i$ is the standard deviation of the
four sub-dimension scores per image
(\S\ref{sec:method:softlabel}); higher $\delta_i$ indicates
greater dimensional disagreement. Tertile boundaries chosen as
described in App.~\ref{app:tertile}, yielding
$n = 468 / 218 / 249$. \textbf{Bold} marks the best score per
column across all $11$ methods. $^{\dagger}$ single-seed result;
remaining ablation rows are 3-seed ensembles. $^{\ddagger}$
diagnostic stack on top of our final composition.}
\label{tab:delta_stratified_full}
\setlength{\tabcolsep}{5pt}
\begin{tabular}{l l c c c c}
\toprule
& \textbf{Method} & low-$\delta$ & mid-$\delta$ & high-$\delta$ & overall \\
&                 & ($n{=}468$)  & ($n{=}218$)  & ($n{=}249$)   & ($n{=}935$) \\
\midrule
\multirow{3}{*}{Handcrafted}
& Qabf                        & 0.353 & 0.269 & 0.180 & 0.344 \\
& VIF                         & 0.352 & 0.366 & 0.197 & 0.340 \\
& MI                          & 0.287 & 0.186 & 0.146 & 0.256 \\
\midrule
\multirow{2}{*}{Zero-shot}
& Q-Align                     & 0.137 & 0.121 & 0.196 & 0.281 \\
& DeQA-Score                  & 0.174 & 0.100 & 0.140 & 0.308 \\
\midrule
In-domain
& ViT-L+MLP                   & 0.432 & 0.426 & 0.194 & 0.554 \\
\midrule
\multirow{5}{*}{\shortstack[l]{Loss-comp.\\ablation}}
& $\sigma$-fixed (DeQA recipe)                                  & 0.566 & 0.597 & \textbf{0.254} & 0.633 \\
& $\sigma$-conflict only                                        & 0.571 & 0.601 & 0.242 & 0.634 \\
& $\sigma$-conflict $+$ $\sigma$-blind PL$^{\dagger}$           & 0.592 & 0.611 & 0.219 & 0.640 \\
& \textbf{Ours} ($\sigma$-conflict $+\,\mathcal{L}_{\mathrm{xfid}}$) & 0.573 & \textbf{0.624} & 0.206 & \textbf{0.648} \\
& Ours $+$ $\sigma$-blind PL$^{\dagger\,\ddagger}$              & \textbf{0.599} & 0.612 & 0.243 & 0.643 \\
\bottomrule
\end{tabular}
\end{table}

\subsection{Tertile Boundary Choice}
\label{app:tertile}

The tertile boundaries used throughout the
$\delta$-stratification analysis ($\delta \le 0.45$,
$0.45 < \delta \le 0.71$, $\delta > 0.71$) are computed once on
the test set and reused across all rows of
Tab.~\ref{tab:delta_stratified_full}. The low boundary is
$0.45$ rather than the smallest possible non-zero std among
integer sub-dim scores ($0.433$) so that images with one
sub-dimension differing by one level land in the low tertile,
yielding the reported $n = 468 / 218 / 249$ split. We verified
that perturbing the low boundary by $\pm 0.02$ shifts $\le 8$
images between low and mid tertiles and does not change any
qualitative claim in Sec.~\ref{sec:exp:delta_validation}.

\subsection{Counterfactual SRCC Ceiling}
\label{app:ceiling}

The residual SRCC headroom not closed by our method is not
entirely a model-side limitation: the high-$\delta$ tertile's
contribution is bounded above by the internal consistency of
the human annotation itself. To quantify how much of the gap is
bridgeable, we compute a counterfactual ceiling that holds the
high-$\delta$ tertile at its empirical SRCC floor (the
within-tertile rater-noise plateau, $\approx 0.21$ across every
cell of our grid) and replaces the low- and mid-$\delta$
within-tertile predictions with their respective oracle ranks
before re-pooling SRCC over the full $935$-image test set. This
places the achievable SRCC on EVAFusion test at approximately
$0.78$, against which our model's $0.6477$ closes a substantial
fraction of the bridgeable headroom over the strongest external
baseline (ViT-L+MLP, $0.554$).

\subsection{Alternative Listwise Scope: PL on the Level
Distribution}
\label{app:pl_dist}

A natural alternative to placing the cross-source-pair
Plackett--Luce listwise term on the scalar $\hat\mu$ readout is
to apply it directly to the $5$-way predicted level distribution
$p \in \Delta^4$. We tested three weights for this variant
($\lambda_{\mathrm{PL\text{-}dist}} \in \{0.05, 0.1, 0.5\}$)
within our training pipeline and observed a consistent
rank-vs-KL trap. The listwise gradient on the level distribution
pulls $p$ toward shapes that maximise pairwise rank ordering,
which conflicts directly with the soft-Gaussian KL supervision
(\S\ref{sec:method:objective}) and substantially inflates the
distribution-prediction loss. At
$\lambda_{\mathrm{PL\text{-}dist}} = 0.5$ -- a weight at which
the scalar-readout listwise term already converges cleanly --
SRCC drops by $0.035$ from the corresponding no-listwise
reference, mean KL inflates by $\sim 90\%$, and ECE rises by a
factor of three. Lowering the weight to $0.1$ or $0.05$ recovers
SRCC to within $\pm 0.005$ of that reference but yields no
positive ranking gain. The cross-source-pair listwise term must
therefore operate on the scalar $\hat\mu$ readout to avoid
degrading the underlying distribution-prediction target -- a
structural constraint that motivates the formulation in
Sec.~\ref{sec:method:objective}.

\begin{table}[h]
\centering
\small
\caption{One-at-a-time $\lambda$-sensitivity sweep around the
submitted recipe (single seed $42$, $n = 935$ test images).
L2 is the reference cell and reuses the published seed $42$
checkpoint of Tab.~\ref{tab:main_results}; L1, L3, L4, L5 are
independently retrained with the indicated weight changed.
Higher is better for all metrics.}
\label{tab:lambda_sweep}
\setlength{\tabcolsep}{4pt}
\begin{tabular}{l c c c c c c c}
\toprule
Cell & $\lambda_{\mathrm{fid}}$ & $\lambda_{\mathrm{xfid}}$ & SRCC$\uparrow$ & PLCC$\uparrow$ & KRCC$\uparrow$ & PairAcc$\uparrow$ & mean $\tau\uparrow$ \\
\midrule
L1                & 0.5 & 0.5  & 0.6234 & 0.6264 & 0.4834 & 0.7635 & 0.4489 \\
\textbf{L2 (ref)} & \textbf{1.0} & \textbf{0.5}  & \textbf{0.6366} & \textbf{0.6395} & \textbf{0.4939} & 0.7665 & 0.4511 \\
L3                & 2.0 & 0.5  & 0.6257 & 0.6292 & 0.4839 & \textbf{0.7770} & \textbf{0.4699} \\
L4                & 1.0 & 0.25 & 0.6332 & 0.6369 & 0.4899 & 0.7677 & 0.4533 \\
L5                & 1.0 & 1.0  & 0.6324 & 0.6349 & 0.4915 & 0.7698 & 0.4572 \\
\bottomrule
\end{tabular}
\end{table}

\subsection{Hyperparameter Sensitivity Sweep}
\label{app:lambda}

We probe whether the submitted recipe
($\lambda_{\mathrm{fid}}\!=\!1.0$,
$\lambda_{\mathrm{xfid}}\!=\!0.5$, referenced in
Sec.~\ref{sec:ablation}) sits at a local optimum or merely a
plateau by sweeping each weight one-at-a-time around the
reference cell while holding all other hyperparameters fixed
(single seed $42$). Tab.~\ref{tab:lambda_sweep} reports the five
resulting cells; the reference (L2) reuses the published seed
$42$ checkpoint without re-training, while L1, L3, L4, L5 are
independently retrained with the indicated weight changed. The
reference attains the highest SRCC ($0.6366$); the four
neighbour cells regress by $-0.0034$ to $-0.0132$ SRCC, with
$\lambda_{\mathrm{fid}}$ exhibiting roughly $3\!\times$ the
sensitivity of $\lambda_{\mathrm{xfid}}$ (half-/double-step
deltas $-0.0132 / -0.0109$ vs $-0.0034 / -0.0042$). The
submitted hyperparameters are therefore at a local optimum on
the $\lambda$ axis under our single-dataset training protocol,
and not a coincidental plateau pick.

\begin{figure*}[t]
\centering
\includegraphics[width=\linewidth]{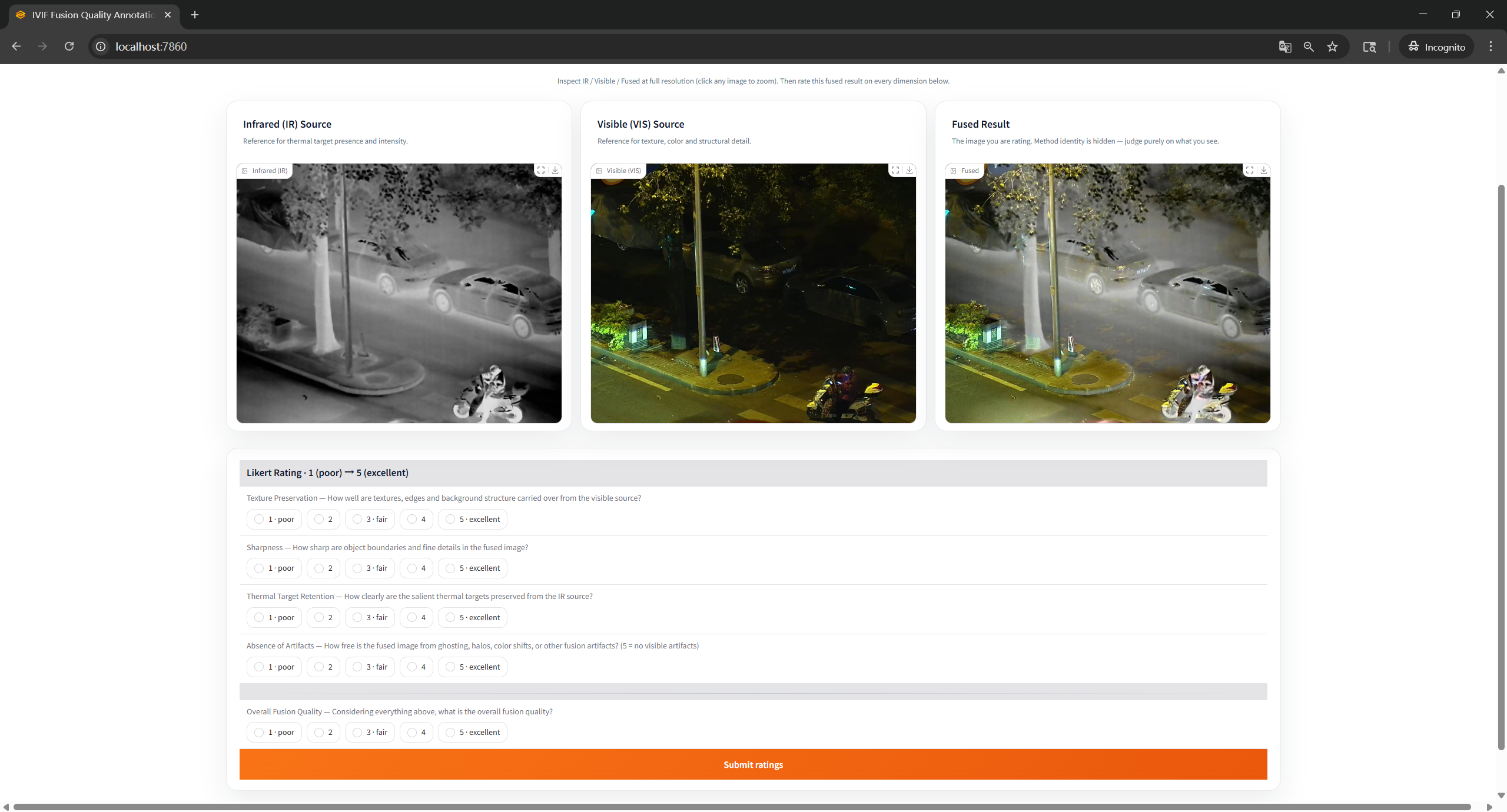}
\vspace{-12pt}
\caption{Human expert annotation web tool.}
\label{fig:quali}
\vspace{-12pt}
\end{figure*}

\section{Expert Annotation Protocol}
\label{app:expert_protocol}

This appendix details the annotation protocol for the
SemanticRT~\cite{ji2023semanticrt} 5-expert pilot benchmark used in
Sec.~\ref{sec:cross_dataset} and the human-side validation
of $\delta_i$ in Sec.~\ref{sec:exp:delta_validation}.

\subsection{Image Selection}

We sample $100$ source pairs from the public SemanticRT test split, ensuring coverage of
the dataset's primary scene categories: outdoor urban,
indoor, low-light, and adverse-weather scenes. Each source
pair is processed by all $11$ fusion methods used in
EVAFusion's evaluation, yielding $1100$ fused images. Method
identity is stripped from filenames before annotation to
prevent method-aware bias.

\subsection{Expert Selection}

Five domain experts were recruited, each meeting the
following criteria: (i) a graduate-level background in
computer vision or image processing, (ii) at least $\geq 2$
prior peer-reviewed publications on image fusion or related
low-level vision tasks, (iii) familiarity with both infrared
and visible imaging modalities. No expert had prior exposure
to the specific Semantic RT images or to FuScore. Experts
were compensated at standard rates per local institutional
guidelines.

\subsection{Rating Protocol}

Each expert independently rated all $1100$ fused images on
both: (i) the four IVIF-specific sub-dimensions (thermal
retention, texture preservation, artifacts, sharpness), and
(ii) the overall fusion quality. All ratings used the same
integer $1$--$5$ Likert scale as EVAFusion. Ratings were collected through a custom web interface that displayed the IR source, visible source, and fused image side-by-side at full resolution, with the ability to zoom into any region.

To minimize systematic biases: (i) the presentation order of
fused images was randomised independently per expert, (ii)
the four sub-dimension prompts were presented in randomised
order per image, (iii) experts could not view or revise
prior ratings during the session, and (iv) all annotations
were completed across $2$--$3$ sessions per expert with
mandatory breaks to mitigate fatigue.

\subsection{Calibration and Quality Control}

Prior to formal annotation, each expert completed a
calibration round on $10$ held-out fused images (not
included in the final $1100$) drawn from a different
source-pair distribution, with rubric-anchored example
images for each quality level. We computed inter-rater
Kendall's $W = 0.65$ on the final $1100$ overall scores,
indicating moderate-to-strong agreement consistent with established IVIF subjective benchmarks~\cite{liu2026bridging}.

\subsection{Ground Truth Aggregation}

The per-image GT for each metric is the arithmetic mean of
the five expert ratings. The per-image inter-rater
disagreement, used in Sec.~\ref{sec:exp:delta_validation},
is the standard deviation of the five overall ratings,
$\mathrm{std}_5(y)$.

\end{document}